\documentclass[]{elsarticle}
\usepackage[a4paper, total={6in, 8in}, margin=0.9in]{geometry}
\usepackage{multicol}
\usepackage{amssymb}
\usepackage{graphicx}
\usepackage{booktabs}
\usepackage{multirow}
 
\usepackage{amsmath}
\usepackage{caption}
\usepackage{subcaption}
\usepackage{algorithm}
\usepackage{algpseudocode}
\usepackage[colorinlistoftodos]{todonotes}
\usepackage{amsmath,amssymb,amsfonts}
\usepackage{algorithm}
\usepackage{algpseudocode}
\usepackage{graphicx}
\usepackage{textcomp}
\usepackage{multirow}
\usepackage{longtable}
\usepackage{rotating} 
\usepackage{longtable}
\usepackage{booktabs}
\usepackage{ltxtable} 
\usepackage{supertabular,booktabs}
\usepackage[export]{adjustbox}
\usepackage{tablefootnote}
\usepackage{color}
\usepackage{xcolor}
\usepackage{url}

\usepackage{changes}
\begin{document}

\begin{frontmatter}

%% Title, authors and addresses

%% use the tnoteref command within \title for footnotes;
%% use the tnotetext command for theassociated footnote;
%% use the fnref command within \author or \address for footnotes;
%% use the fntext command for theassociated footnote;
%% use the corref command within \author for corresponding author footnotes;
%% use the cortext command for theassociated footnote;
%% use the ead command for the email address,
%% and the form \ead[url] for the home page:
%% \title{Title\tnoteref{label1}}
%% \tnotetext[label1]{}
%% \author{Name\corref{cor1}\fnref{label2}}
%% \ead{email address}
%% \ead[url]{home page}
%% \fntext[label2]{}
%% \cortext[cor1]{}
%% \affiliation{organization={},
%%             addressline={},
%%             city={},
%%             postcode={},
%%             state={},
%%             country={}}
%% \fntext[label3]{}

\title{A Differentially Private Federated Proximal Optimization Framework for Customer Churn Prediction in Heterogeneous Federated Telecom Networks}

%% AUTHORS %%%%%%%%%%%%%%%%%%%%%%%%%%%%%%%%%%%%%%%%%%%%%%%%%%%%%%%%%%%%%%%%%%%%
%% Leave this section commented out so that the paper is blinded for review.
%% Group authors per affiliation:

%% AUTHORS %%%%%%%%%%%%%%%%%%%%%%%%%%%%%%%%%%%%%%%%%%%%%%%%%%%%%%%%%%%%%%%%%%%%
%% Leave this section commented out so that the paper is blinded for review.
%% Group authors per affiliation:
\author[ss]{Joydeb Kumar Sana\corref{cor1}}
\address[ss]{Institute of Information and Communication Technology, Bangladesh University of Engineering and Technology, Dhaka, Bangladesh (e-mail: joysana@gmail.com, joydebsana@iict.buet.ac.bd)}

\author[bb]{Subrata Chakraborty}
\address[bb]{School of Science and Technology, Faculty of Science, Agriculture, Business and Law, University of New England, Armidale, NSW, Australia (e-mail: Subrata.Chakraborty@une.edu.au)}

\author[mm]{M M Manjurul Islam}
\address[mm]{Intelligent Systems Research Centre, Ulster University, Londonderry BT48 7JL, UK  (e-mail: m.islam@ulster.ac.uk)}

%% Only give the email address of the corresponding author
\cortext[cor1]{Corresponding author (Joydeb Kumar Sana)}

%%%%%%%%%%%%%%%%%%%%%%%%%%%%%%%%%%%%%%%%%%%%%%%%%%%%%%%%%%%%%%%%%%%%%%%%%%%%%%%%

%% use optional labels to link authors explicitly to addresses:
%% \author[label1,label2]{}
%% \affiliation[label1]{organization={},
%%             addressline={},
%%             city={},
%%             postcode={},
%%             state={},
%%             country={}}
%%
%% \affiliation[label2]{organization={},
%%             addressline={},
%%             city={},
%%             postcode={},
%%             state={},
%%             country={}}

%\author{}

%\affiliation{organization={},%Department and Organization
%            addressline={}, 
 %           city={},
  %          postcode={}, 
   %         state={},
    %        country={}}

%\author{  }

\begin{abstract}
Customer churn is one of the major issues in the telecommunication industry. To predict customer churn, conventional centralized machine learning approaches have been widely used. This centralized approach requires customer data to be stored in a central repository, which raises privacy concerns and may violate data protection regulations. Federated learning addresses this problem by allowing multiple telecom operators to collaboratively train a global model without transferring their raw customer data. However, real-world customer data are often heterogeneous (non-IID), which may negatively affect the performance of standard federated learning. Trained models can also suffer from privacy attacks. To address those issues, we propose a Differentially Private (DP) based Federated Proximal optimization (FedProx) framework. All experiments were performed on two publicly available telecom churn datasets. We trained Federated Averaging (FedAvg), DP-FedAvg, FedProx, and the proposed DP-FedProx framework. For baseline comparison, we also used several centralized and local models. To evaluate the models, we employed seven widely used evaluation metrics. The experimental results show that the FedProx based models consistently outperform the FedAvg based models. Compared with the best centralized model, the proposed DP-FedProx framework achieves competitive prediction performance with only a small reduction in accuracy while providing privacy guarantees. To explain our model, we conducted SHAP analysis which shows that DP-FedProx method priorities revenue group features. These results indicate that the proposed DP-FedProx framework provides a practical balance between prediction performance and data privacy protection.
\end{abstract}

\begin{keyword}
Data Privacy, Diffential Privacy, Federated Learning, FedAvg, FedProx,  Distributed Machine Learning, Customer Churn, Telecommunication Industry
%% keywords here, in the form: keyword \sep keyword
%% PACS codes here, in the form: \PACS code \sep code
%% MSC codes here, in the form: \MSC code \sep code
%% or \MSC[2008] code \sep code (2000 is the default)
\end{keyword}

\end{frontmatter}

\section{Introduction}

Customers are the most important part of every business. Businesses generally depend on customer satisfaction. In this competitive market, customers have numerous choices. They can switch service providers. These customers are classified as churned customers\cite{JDK_2025_PPCCP, JDK_2022_CCP}. This customer churn phenomenon is frequently observed in the telecommunications industry.  It is a serious issue because the revenue of the service provider is highly dependent on the retention of existing customers.  Customer churn affects the organizational profitability, market share,  long-term sustainability, fame and branding. Previous studies suggest that retaining existing customers is considerably less costly than acquiring new ones \cite{JDK_2025_PPCCP, JDK_2022_CCP, Saha2024}. Therefore, customer retention remains one of the most important objectives in the telecommunications industry.  In this competitive market,  it is essential for the service providers to identify the customers who are going to churn. If a  service provider can prediction the customer whos are likey to churn, it can cater targeted offerings to them to reduce their dissatisfaction.

To predict the churned customer, over the few years, machine learning based methodologies have attracted  significant attention both academia and industry \cite{Lalwani2021, Chang2024}. However, traditional customer churn prediction models rely on centralized machine learning frameworks in which customer data from multiple branches, service regions, or organizations are aggregated into a central repository for model training \cite{Chen2024}. While centralized approach achieve high predictive performance, it often introduce significant privacy and regulatory concerns \cite{Chen2024}. Usually, it does not meet the regulatories such as General Data Protection Regulation (GDPR) and other data governance frameworks \cite{Huh2024}. As a result, telecommunication companies are increasingly seeking privacy-preserving machine learning solutions that does not require transforming customer data from branches. To address this issue researchers use distributed Federated Learning (FL) \cite{McMahan2016, Yin2021}. Instead of transferring customer data to a centralized server, FL learning approache is capable to training a shared global model keeping the customer data to its branch. As this technique does not need data sharing it reduces privacy risks significantly. There are several FL approches such as Federated Averaging (FedAvg), FedAdam, FedYogi, SCAFFOLD (Stochastic Controlled Averaging),  etc. \cite{Zantalis2026, Reddi2020, Karimireddy2019}. Among them FedAvg has become the most widely adopted aggregation algorithm due to its simplicity and effectiveness \cite{Li2020, Sahu2018, McMahan2016}. However, it assume that client data are independently and identically distributed (IID) which is rare in real-world telecommunications environments \cite{Zhao2018, Li2020, Lim2020}. To overcome this limitations for heterogeneous environments researchers introduce FedProx aggregation technique which is a proximal regularization \cite{prathusha2025}. This technique has the potential to improved robustness and convergence behavior in non-independent and identically distributed (non-IID) settings.

Though FL eliminates the data sharing,  recent studies have revealed that sensitive information may still be inferred from exchanged model updates through gradient inversion and reconstruction attacks \cite{Jiang2025, Huang2021}. Moreover, trained model may suffer privacy attacks \cite{JDK_2025_PPCCP}. This remains a major privacy concern during the training process and for the trained models. In this context, differential privacy (DP) has emerged as a promising approach to overcome this problem \cite{JDK_2025_PPCCP}. To the best of our knowledge, despite significant advancements in FL and DP techniques, no research has investigated the combined impact of non-IID data based FL strategy and DP on customer churn prediction for the telecommunication industry. In this research, we propose a DP based  federated learning framework for customer churn prediction in the telecommunication industry that addresses both statistical heterogeneity and privacy concerns.

\subsection{Literature Review} \label{sec:literature_review} 

%\subsection{Customer Churn Prediction in Telecommunications}
Over the few years, customer churn problem is the one of the major challenges in the telecom industry. To tackle this problem, several machine learning approaches have been utilized such as   Logistic regression\cite{JDK_2025_PPCCP, JDK_2022_CCP}, Decision Trees \cite{JDK_2025_PPCCP, JDK_2022_CCP}, Random Forests \cite{JDK_2025_PPCCP}, Support Vector Machines, Gradient Boosting \cite{JDK_2025_PPCCP}, Extreme Gradient Boosting (XGBoost), and Artificial Neural Networks etc. Researchers in \cite{Shaikhsurab2024, Liu2022} employed deep, hybrid, and ensemble learning models to better capture the nonlinear behavior of customers. Amin et al. in \cite{Amin2023}, used adaptive Naïve Bayes approach with evolutionary feature weighting and achives notable performnace improvement. Telecom churn datasets are often class-imbalanced and high-dimensional, which can bias predictions and reduce generalizability \cite{Shaikhsurab2024}. Several studies therefore stress preprocessing, resampling, feature extraction, and clustering or segmentation as central to performance rather than treating classifier choice alone as decisive \cite{Liu2022, Ouf2024}. Ensemble learning methods \cite{Adiputra2023}, adpative Ensemble learning \cite{Shaikhsurab2024}, and deep learning architectures have consistently demonstrated superior performance due to their capability to model intricate customer behavior patterns and interactions among features. Despite their predictive success, most existing churn prediction frameworks assume centralized access to customer data, which is often impractical in privacy-sensitive environments.

%\subsection{Federated Learning}
Training a model using a centralized data is a privacy concern. Organizations are reluctant to share sensitive customer records across firms or cloud environments\cite{JDK_2025_PPCCP}. Federated learning addresses this by keeping data local while training a shared model, making it suitable for churn prediction across providers \cite{Yin2025, Fahmy2025}.  Hug et al. in \cite{Huh2024}, used FL for churn prediction in Telecom industry. However, it did not use the full potential of FL. Banking churn study found that federated learning improved prediction accuracy by about 2\% relative to local-only training while preserving privacy \cite{Fahmy2024}. Okonkwo et al. in \cite{Okonkwo2025} argued that federated learning can improve centralized predictive performance while preserving privacy. In \cite{JDK_2022_CCP}, Sana et al. employed a privacy preserving framework combining with GAN and adaptive weight-of-evidence method that generated synthetic telecom data. This methodology achieved 87.1\% F-measure and gains up to 28.9\% in accuracy and 27.9\% in F-measure over earlier methods. This shows that privacy protection in churn prediction has been pursued through both distributed learning and privacy-preserving data transformation, but no common framework currently exists to manage heterogeneous data effectively. Federated Learning was introduced as a distributed machine learning paradigm that enables collaborative model training without requiring the exchange of raw data\cite{Liu2021, McMahan2016, Bonawitz2021}. This approach significantly reduces privacy risks while enabling knowledge sharing across distributed data sources. 
Although federated learning protects raw data from direct exposure, recent research has revealed that sensitive information may still be inferred from trained models through gradient inversion attacks, membership inference attacks, and reconstruction attacks \cite{JDK_2025_PPCCP}. To overcome form this privacy concern, differential Privacy has emerged as one of the most widely adopted privacy-preserving techniques in machine learning. Differentially Private Stochastic Gradient Descent (DP-SGD) is one of the widly used DP method that inject noises during the trainng process, ensuring privacy guarantee. To the best of our knowledge, only two studies \cite{Huh2024, Krishnan2024} have focused on FL based customer churn prediction. However, the study \cite{Huh2024} compromises the core privacy objective of federated learning by transferring raw customer data to a centralized global repository. On the other hand, the study by~\cite{Krishnan2024} employed the FedAvg algorithm but did not incorporate differential privacy. Moreover, the effectiveness of DP-based FedProx federated learning for customer churn prediction in the telecommunications industry has not yet been studied. This research addresses these gaps by proposing and evaluating a privacy-preserving DP-FedProx framework for customer churn prediction under non-IID data distributions.

\subsection{Our Contributions}

The main contributions of this work are summarized as follows:

\begin{itemize}

\item We propose a privacy-preserving federated learning framework for telecom customer churn prediction without sharing raw customer data.

\item We integrate the FedProx optimization algorithm with Differential Privacy (DP) to address two major challenges in federated churn prediction: statistical heterogeneity and privacy leakage.

\item We simulate a realistic federated telecom environment using tenure/ month based client partitions to capture data heterogeneity.

\item The performance of centralized, local, standard FedAvg, standard FedProx, DP-based FedAvg, and DP-based FedProx models was systematically evaluated.
 
\item We provide both global and client-level performance assessments on two publicly available benchmark datasets using seven evaluation metrics: Accuracy, Precision, Recall, F1-score, Specificity, ROC-AUC, and PR-AUC.

%\item Experimental results demonstrate that the proposed FedProx-DP framework achieves strong predictive performance while preserving privacy under non-IID settings.

\item To the best of our knowledge, this study is the first to incorporate Differential Privacy into a federated learning for the customer churn prediction in the telecom industry.

\end{itemize}

\section{Methodology}
\label{sec:methodology}

This section presents the overall design adopted for the federated customer churn prediction model under non-IID data distributions and differential privacy constraints.  We describe the dataset, the non-IID partitioning strategy, data processing, the DP mechanism, the neural network architecture shared by all federated experiments, and the evaluation measures.

\subsection{Dataset and Non-IID Client Partitioning}
\label{subsec:data-partitioning}

In this research, we have utilized two publicly available telecom customer churn datasets. Each record contains demographic, account, and usage-based features along with a binary churn label. Rows with missing values are removed, and the customer identifier column is dropped as a non-predictive key. The churn label is normalized to a binary $\{0, 1\}$ encoding, and categorical features are label-encoded prior to client partitioning. The sample sizes of the dataset-1 and  dataset-2 are $100000$ and  $7043$, respectively. More details about these datasets are presented in Table \ref{table:dataset}.

% ===== NEW: geographic provenance paragraph, answers reviewer Q1 =====
%Both datasets originate from the U.S. telecommunications market, though with differing levels of documented provenance. 
Dataset-2 is IBM's published sample dataset for Cognos Analytics/Watson, explicitly documented as covering a fictional carrier providing home phone and Internet service to 7043 customers in California, USA, during a fiscal quarter \cite{IBM_Telco_Docs}. Dataset-1 does not carry equivalent official documentation on Kaggle. The schema and scale (records/usage/billing/demographic feature families) of dataset-1 are consistent with the widely-cited Cell2Cell dataset released by the Teradata Center for Customer Relationship Management at Duke University \cite{teradata2002cell2cell}. We note this lineage as inferred from structural similarity rather than as a confirmed citation chain, since the uploader's Kaggle listing does not itself state a source. 
% ===== END NEW =====

\begin{table}[h!]
\caption{Summary of datasets}
%\label{table:1}
\label{table:dataset}
\begin{center}
\begin{tabular}{ p{5cm} p{2cm}  p{2cm}  }
 \hline
 \vspace{.05mm}\\
 Description& Dataset-1 &Dataset-2 \\
 \vspace{.05mm}\\
 \hline
 \vspace{.01mm}\\
 No. of samples   & 100000    &7043 \\
 No. of attributes&   101  & 21  \\
 No. of class labels & 2 & 2 \\
 Percentage of churn samples    & 49.56 & 26.54  \\
 Percentage of non-churn samples &  50.43  & 73.46 \\
 Source of the datasets& Kaggle \cite{Dataset_1_2022}     & Kaggle \cite{Dataset_2_2022}    \\
 \hline
\end{tabular}
\end{center}

%URL$^1$: https://www.kaggle.com/abhinav89/telecom-customer/data (Last Access: February 15, 2022).\\
%URL$^2$:https://www.kaggle.com/blastchar/telco-customer-churn (Last Access: February 15, 2022).\\

\end{table}

To simulate a realistic non-IID federated setting, we partition customers into four clients based on account tenure (in months), reflecting natural heterogeneity across customer lifecycle segments rather than an artificial or uniformly random split:

\begin{itemize}
    \item \textbf{Client 1 -- Short tenure:} tenure $\leq 12$ months
    \item \textbf{Client 2 -- Mid tenure:} $12 <$ tenure $\leq 24$ months
    \item \textbf{Client 3 -- Long tenure:} $24 <$ tenure $\leq 48$ months
    \item \textbf{Client 4 -- Loyal tenure:} tenure $> 48$ months
\end{itemize}

This partitioning creates differences in both the feature values and class distributions across clients ( e.g. customer usage patterns, contract types, churn rates) with customer tenure. It provides a realistic way to test how well federated learning methods perform when client data are heterogeneous. This tenure-based partitioning is applied independently within each of the two datasets, so no client is ever formed by mixing records from Dataset-1 and Dataset-2, and the resulting federated experiments for each dataset are evaluated on their own.

\subsection{Adaptive Weight-of-Evidence Feature Encoding}
\label{subsec:woe-encoding}

To handle skewed and high-cardinality numerical features in a manner that captures their relationship with the target variable while remaining computationally lightweight for downstream neural and tree-based models, we apply a custom Weight-of-Evidence (WoE) \cite{JDK_2025_PPCCP} encoding scheme to each client's feature set \emph{independently}, preventing cross-client information leakage during preprocessing.

For a feature column $x$, the encoder first determines whether the cardinality of $x$ exceeds 100 unique values. If so, $x$ is discretized into $B$ quantile bins, where
\begin{equation}
B = \left\lfloor \frac{n_k}{10} \right\rfloor,
\end{equation}
with $n_k$ the number of samples at client $k$, adapting bin granularity to local sample size. Low-cardinality features are used directly as categorical groups without binning.

For each resulting bin or category $g$, the WoE value is computed as

\begin{equation}
\text{WoE}(g) = \ln \left( \frac{P(g \mid y=1)}{P(g \mid y=0)} \right) = \ln \left( \frac{\text{pos}_g / \sum \text{pos}}{\text{neg}_g / \sum \text{neg}} \right),
\end{equation}

where $\text{pos}_g$ and $\text{neg}_g$ are the counts of churned and non-churned customers, respectively, falling into group $g$. Groups with fewer than a minimum sample threshold, or with zero positive or zero negative instances, are assigned a neutral WoE value of $0$ to avoid degenerate log-ratios. Missing values are imputed with a sentinel value ($-9999$) prior to binning so that missingness itself can carry predictive signal. Each feature column is then replaced by its corresponding WoE value, yielding a fully numeric, monotonically informative representation of the original feature with respect to the churn outcome.

\subsection{Adaptive Weight-of-Evidence Feature Encoding}
\label{subsec:woe-encoding}

To transform the input features into a more informative numerical representation, we apply a custom Weight-of-Evidence (WoE) encoding method \cite{JDK_2025_PPCCP} to each client's data independently. This encoding process performed separately for each client during data preprocessing. For each feature, we first check the number of unique values. If a feature has more than 100 unique values, it is divided into quantile-based bins. The number of bins is determined by

\begin{equation}
B = \left\lfloor \frac{n_k}{10} \right\rfloor,
\end{equation}

where $n_k$ is the number of samples at client $k$. This allows the number of bins to adapt to the size of the client's dataset. Features with 100 or fewer unique values are used directly as categorical groups without binning.

For each bin or category $g$, the WoE value is calculated as

\begin{equation}
\text{WoE}(g) = \ln \left( \frac{P(g \mid y=1)}{P(g \mid y=0)} \right)
= \ln \left( \frac{\text{pos}_g / \sum \text{pos}}{\text{neg}_g / \sum \text{neg}} \right),
\end{equation}

where $\text{pos}_g$ and $\text{neg}_g$ are the numbers of churned and non-churned customers in group $g$, respectively. Groups with fewer than a minimum sample threshold, or with zero positive or zero negative instances, are assigned a neutral WoE value of $0$ to avoid degenerate log-ratios. Finally, each original feature value is replaced with its corresponding WoE value.

\subsection{ Federated Learning, FedAvg, and FedProx}
\label{subsec:fl-background}

\subsubsection{Federated Learning}
Federated Learning (FL) is a distributed machine learning approach where multiple clients collaterally train a shared global model. Instead of storing each parties data to a central server, each client trains the modle localy on its own data and shared the parameter values \cite{McMahan2016}. A server aggregates these updates and improves the global model. The server then redistributes the updated model  to clients for the next round of local training. After several iterative rounds the global model  converges. 

FL was developed to tackle the privacy concern of the sensitive data where data owners are reluctant to share their data to a central storage to train a machine learning model. Telecom operators often do not want to store their customer information in a centralized repository due to privacy regulations, security risks, competitive concerns, and data governance requirements \citep{McMahan2016, Kairouz_2021}. In this context, FL can help to overcome those challenges. There are several FL approaches have been used in academia and engineering practice including  FedAvg, FedProx, FedAdam, FedYogi, FedAdagrad, Scaffold, FedNova, etc. \cite{Yurdem2024,Reddi2020,Karimireddy2019,Sahu2018}. Among them FedAvg is widely used technique due to its efficiency and simplicity \cite{McMahan2016,Sahu2018}. FedProx is a lightweight generalization of FedAvg which is well accepted for its robustness and stability in heterogeneous setting. For independently and identically distributed (IID) data, FedAvg works well but for non-independent and identically distributed (non-IID) data, it degrades the quality of the aggregated global model and slows down convergence \citep{Li2020}.  This challenge motivates us to test the effect of those two approaches for the customer churn prediction in the telecom industry.

%\subsubsection{Federated Averaging (FedAvg)}
Federated Averaging (FedAvg) first introduced by McMahan et al. in \cite{McMahan2016}. In this FL algorithm, the global model converges with several communication round. At first, the server broadcasts the current global model parameters and each clients initializes its local model with the received global parameters. The clients train their local models using their own data and sent their updated model parameter values back to the server. The server then averages of the received values and forms a new global model. Typically, it averages the parameter values weighted by the number of training samples held by each client. Though FedAvg, substantially reduce communication overhead, under non-IID data, it shows slow convergence and degrade final model quality \citep{Li2020}.

%\subsubsection{Federated Proximal Optimization (FedProx)}
Federated Proximal optimization (FedProx) was proposed by Li et al. in \cite{Li2020}. It is a extension of FedAvg which improves the robustness in systems heterogeneity. It also uses weighted averaging to aggregate client updates but modifies the local objective optimized at each client. FedProx uses proximal regularization as loss function to each client. This constraint keeps the local model closer to the global model during the training process. As a result, it improves training stability and performance \cite{Li2020}. This approach uses a proximal coefficient $\mu$ which controls the FedProx behavior (as $\mu \to 0$, the proximal term vanishes and FedProx reduces to FedAvg). When the value of $\mu$ increases, the local models are forced to stay closer to the global model. This improves training stability and model performance for the non-IID data distribution \cite{Li2020}. To test the effectiveness of these features of FedProx, we used this method in this study.

\subsection{Differential Privacy}
\label{subsec:dp-background}
Differential Privacy (DP) is first introduced by Cynthia Dwork in \cite{DworkCynthia2006}. It is a mathematical framework that protect data privacy. To protect data privay, it inject random noise into raw data without jeopardizing the data quality and distribution. Informally, a randomized mechanism is differentially private if the outcome of any query executed on the modified dataset and on the real dataset must not be distinguishable.  This property ensures the data privacy from an adversary. Formally, a randomized mechanism $\mathcal{M}$ satisfies $(\epsilon, \delta)$-differential privacy \cite{DworkCynthia2006} if, for any two datasets $D$ and $D'$ differing in exactly one record, and for any measurable output set $S$,
\begin{equation}
\Pr[\mathcal{M}(D) \in S] \leq e^{\epsilon} \, \Pr[\mathcal{M}(D') \in S] + \delta.
\end{equation}
The parameter $\epsilon$ is the privacy budget which controls the the privacy guarantee. The smaller value of $\epsilon$ gives stronger privacy, while larger values provides weaker protection. The parameter $\delta$ represents the probability of a privacy breach.  It dictates how often that protection might completely fail or leak. If $\delta$ is set to exactly $0$, the algorithm is considered to be purely differentially private ($\epsilon$-DP). If delta is greater than $0$, it allows for approximate differential privacy ($\epsilon$, $\delta$-DP), which is often necessary to compute complex machine learning models without destroying data utility \cite{JDK_2026_CAADP}.

There are several DP methods, including noise-addition mechanisms such as the Laplace mechanism \cite{Geng_2016}, Gaussian mechanism \cite{Jinshuo2022}, and Exponential mechanism \cite{Frank2007}, as well as algorithmic frameworks such as Differentially Private Stochastic Gradient Descent (DP-SGD) and Private Aggregation of Teacher Ensembles (PATE) \cite{Jordon2019PATEGANGS} that incorporate these mechanisms.
Among them, the most widely adopted approach is DP-SGD \cite{abadi2016deep} due to its strong privacy guarantees, high momechanismdel utility and suitable to intrigate with standard deep learning approaches. In this work, DP-SGD mechanism has been used in our federated settings. The DP is applied independently at each participating client.

\subsubsection{Centralized, Local, and Federated Learning Classifiers}

For comparison, we trained three types of models. First, we trained several centralized machine learning models on the combined dataset, including Random Forest, Logistic Regression, XGBoost, Gradient Boosting, AdaBoost, and ChurnNet. Second, we trained local models, where each client independently trained a Logistic Regression model using only its own data without collaborating with other clients. The overall performance of the local models was calculated by combining the predictions from all four clients. Third, all federated learning experiments used the same neural network model, called \textit{ChurnNet}, to ensure a fair comparison.

ChurnNet is a feed-forward neural network with two hidden layers and is defined as

\begin{equation}
    f_\theta(x) = \sigma\Big(W_3 \, \phi_2\big(\mathrm{BN}_2(W_2 \, \phi_1(\mathrm{BN}_1(W_1 x)))\big)\Big),
\end{equation}

where $W_1 \in \mathbb{R}^{128 \times d}$ and $W_2 \in \mathbb{R}^{64 \times 128}$ are the weight matrices of the hidden layers, and $\mathrm{BN}_1$ and $\mathrm{BN}_2$ represent batch normalization. The functions $\phi_1$ and $\phi_2$ denote the ReLU activation function followed by dropout with rates of 0.3 and 0.2, respectively. The output layer, represented by $W_3 \in \mathbb{R}^{1 \times 64}$, uses the sigmoid function $\sigma(\cdot)$ to produce the probability of customer churn. The model is trained using the binary cross-entropy loss function.

\subsection{Model Architecture} \label{subsec:model-architecture}
\begin{figure}[!htb]
\centering
\fbox{\includegraphics[height=450px,width=325px]{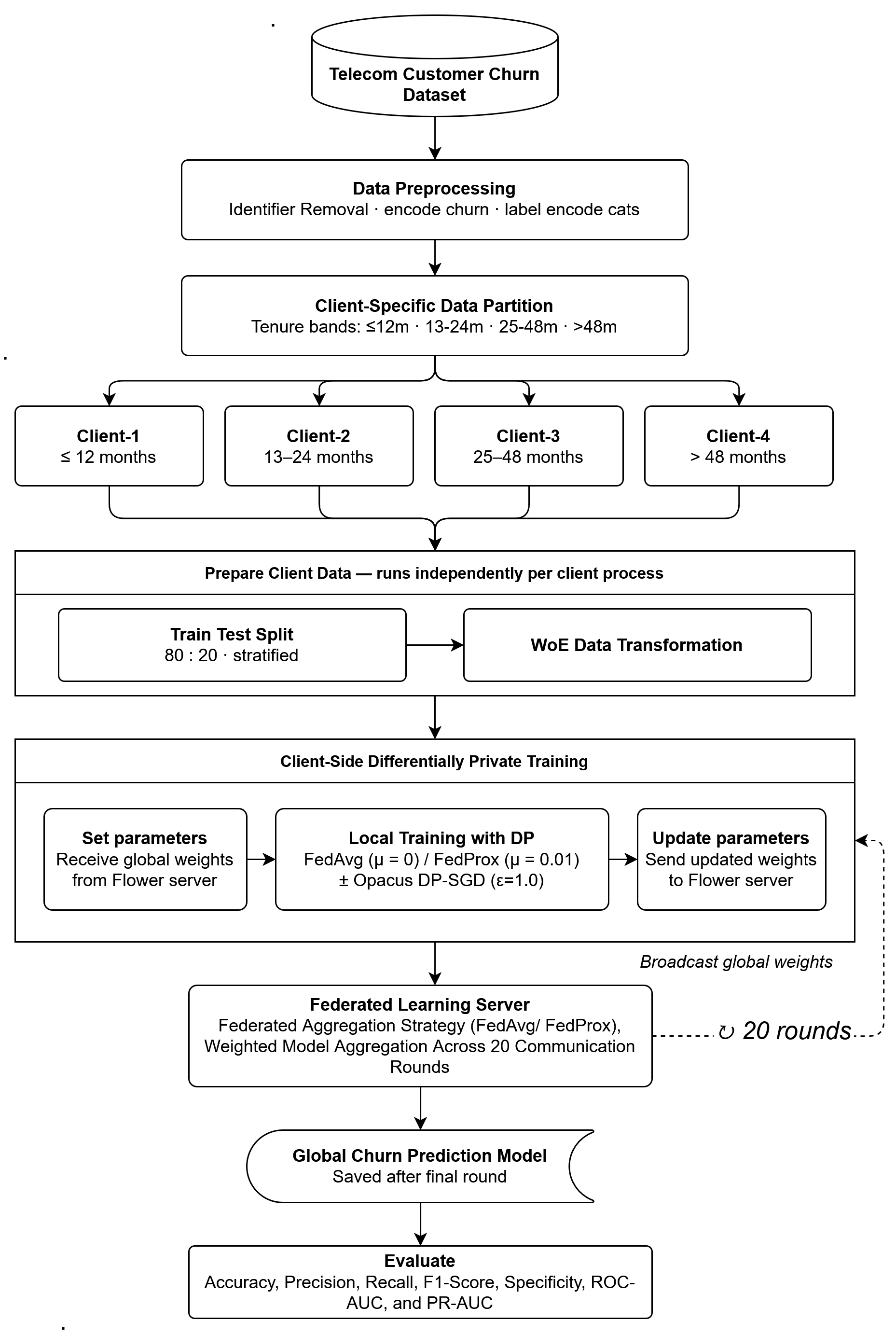}}
\caption{Block Diagram of Differentially Private Federated Learning Framework for Churn Prediction}
\label{fig:DP-FL-churnnet-arch}
\end{figure}

Figure~\ref{fig:DP-FL-churnnet-arch} presents the workflow of the proposed Differentially Private Federated Learning (DP-FL) framework for customer churn prediction. After data preprocessing, we partitioned the data into four non-IID clients based on customer tenure: Client-1 (($\leq$ 12) months), Client-2 (13 - 24 months), Client-3 (25 - 48 months), and Client-4 ((>48) months). Before starting the training process, each client divides its local data into training and testing sets and applies the Weight of Evidence (WoE) transformation on the datasets. During the training, the server sends the current global model parameters to all the clients. Each client performs local training using those parameters. The local training has been performed using both FedAvg ($\mu$=0) and FedProx ($\mu$=0.01) approaches. Differentially Private Stochastic Gradient Descent (DP-SGD) with Opacus has been applied during local training. After performing local training, each client sends back its updated model parameter values to the server and server aggregates them. The same FedAvg or FedProx method has been used in server to update the global model. This client-server training process performs 20 rounds and the global model converges. This final global model is evaluated using seven evaluation metrics. 

The main focus of this research is to develop a privacy preserving model in non-IID data settings, we proposed the FedProx FL based method.  The complete procedure 
is formalised in Algorithm~\ref{alg:proposed}.
 
\begin{algorithm}[H]
\caption{FedProx + DP (Proposed Method)}
\label{alg:proposed}
\begin{algorithmic}[1]
\Require $K$ clients, $T$ rounds, $E$ local epochs, $B$ batch size,
         $\mu$ proximal coefficient, $C$ clip norm,
         $(\varepsilon, \delta)$ privacy budget
\State Initialise global model $\boldsymbol{\theta}^{(0)}$
\For{$t = 1, \ldots, T$}
  \State Server broadcasts $\boldsymbol{\theta}^{(t-1)}$ to all clients
  \For{each client $k \in \{1, \ldots, K\}$ \textbf{in parallel}}
    \State $\boldsymbol{\theta}_k \leftarrow \boldsymbol{\theta}^{(t-1)}$
    \State Attach Opacus \texttt{PrivacyEngine} to
           ($\boldsymbol{\theta}_k$, Adam, $\mathcal{D}_k$)
           with target $(\varepsilon, \delta)$, clip norm $C$
    \For{$e = 1, \ldots, E$}
      \For{each mini-batch $\mathcal{B} \sim \mathcal{D}_k$}
        \State Compute per-sample gradients of
               $\mathcal{L}_{\mathrm{BCE}} +
               \frac{\mu}{2}\|\boldsymbol{\theta}_k - \boldsymbol{\theta}^{(t-1)}\|^2$
        \State Clip: $\tilde{\mathbf{g}}_i \leftarrow
               \mathbf{g}_i / \max(1, \|\mathbf{g}_i\|_2 / C)$
        \State Aggregate and add noise:
               $\hat{\mathbf{g}} \leftarrow \frac{1}{|\mathcal{B}|}
               \bigl(\sum_i \tilde{\mathbf{g}}_i +
               \mathcal{N}(\mathbf{0}, \sigma^2 C^2 \mathbf{I})\bigr)$
        \State Update: $\boldsymbol{\theta}_k \leftarrow
               \text{Adam}(\boldsymbol{\theta}_k, \hat{\mathbf{g}})$
      \EndFor
    \EndFor
    \State Upload $\boldsymbol{\theta}_k^{(t)}$ to server
  \EndFor
  \State $\boldsymbol{\theta}^{(t)} \leftarrow
         \sum_{k=1}^{K} \frac{n_k}{n} \boldsymbol{\theta}_k^{(t)}$
         
\EndFor
\State \Return $\boldsymbol{\theta}^{(T)}$
\end{algorithmic}
\end{algorithm}

\subsection{Evaluation Measures}
\label{subsec:evaluation_measure}
In this study, we used seven widely adopted evaluation metrics. All metrics were calculated based on the confusion matrix, which consists of four prediction outcomes: True Positives (TP), True Negatives (TN), False Positives (FP), and False Negatives (FN). A positive label (1) denotes a churn event, and a negative label (0) denotes non-churn event. The description evaluation metrics are given below.

\begin{description}
  \item[Accuracy:] It is a proportion of correctly classified instances:
   \begin{equation}
      \mathrm{Acc} = \frac{TP + TN}{TP + TN + FP + FN}.
    \end{equation}
 
  \item[Precision:] The positive predictive value, which measures the proportion of predicted churners who actually churned.
    \begin{equation}
      \mathrm{Prec} = \frac{TP}{TP + FP}.
    \end{equation}
 
  \item[Recall (Sensitivity):] It is the true positive rate, which measures the proportion of actual churners that are correctly identified:
    \begin{equation}
      \mathrm{Rec} = \frac{TP}{TP + FN}.
    \end{equation}
 
  \item[F1-Score:] It is the harmonic mean of precision and recall:
    \begin{equation}
      F_1 = 2 \cdot \frac{\mathrm{Prec} \cdot \mathrm{Rec}}
                         {\mathrm{Prec} + \mathrm{Rec}}.
    \end{equation}
 
  \item[Specificity] True negative rate; the fraction of actual non-churners correctly identified:
    \begin{equation}
      \mathrm{Spec} = \frac{TN}{TN + FP}.
    \end{equation}
 
  \item[ROC-AUC:] The area under the Receiver Operating Characteristic (ROC) curve, which measures the model's ability to distinguish between churners and non-churners across all classification thresholds:
    \begin{equation}
      \mathrm{AUC}_{\mathrm{ROC}}
      = \int_0^1 \mathrm{TPR}(t)\, \mathrm{d}[\mathrm{FPR}(t)].
    \end{equation}
 
  \item[PR-AUC:] The area under the Precision-Recall (PR) curve, which evaluates how well the model identifies churners:
    \begin{equation}
      \mathrm{AUC}_{\mathrm{PR}}
      = \int_0^1 \mathrm{Prec}(r)\, \mathrm{d}r.
    \end{equation}
    
\end{description}

\subsection{Coding and Experimental Environment} \label{sec:coding_env}
All experiments were conducted on a Windows Server 2022 (64-bit) machine equipped with an Intel Xeon Silver 4214R processor (2.4 GHz), 64 GB RAM, and 1 TB storage, using Python 3.12. The key libraries used are PyTorch, Opacus (DP-SGD), Flower (federated learning), scikit-learn, XGBoost, etc. The dataset is partitioned into four non-IID client shards based on customer tenure. We split the dataset into 80\% training and 20\% testing sets using stratified sampling. All experiments are run with  a fixed random seed (42) for reproducibility.  Hyperparameter settings are summarised in Table~\ref{tab:hyperparams}. The complete set of code that we used for our experiment  is available at: \url{https://github.com/joysana1/DP-FL-CCP}.  
 
\begin{table}[h]
  \centering
  \caption{Hyperparameter settings for all experiments.}
  \label{tab:hyperparams}
  \begin{tabular}{lll}
    \toprule
    \textbf{Parameter} & \textbf{Value} \\
    \midrule
    Clients $K$                  & 4                \\
    Communication rounds $T$     & 20               \\
    Local epochs $E$             & 5                \\
    Batch size $B$               & 64               \\
    Learning rate $\eta$         & $10^{-3}$        \\
    Proximal coefficient $\mu$   & 0.01             \\
    Clip norm $C$                & 1.0              \\
    Privacy budget $\varepsilon$ & 1.0             \\
    Privacy delta $\delta$       & $10^{-5}$      \\
    RF estimators                & 200             \\
    LR max iterations            & 1{,}000        \\
    Hidden layer sizes           & $[128, 64, 1]$  \\
    Dropout rates                & $[0.3, 0.2]$    \\
    Normalisation                & GroupNorm      \\
    \bottomrule
  \end{tabular}
\end{table}

\section{EXPERIMENTAL RESULTS} \label{sec:experimental_result}
In this section, we presents detailed experimental results. To access the performance of the proposed methodology, we conducted several experiments on two different datasets and utilized seven evaluation metrics. We treated each dataset independently rather than pooling them together. For the differential privacy based federated learning we set the $\epsilon$ value is $1.0$;

\subsection{Results on Dataset-1}

\begin{table*}[!ht]
\centering
\caption{Performance comparison of centralized, local, federated, and differentially private federated learning models on Dataset-1. The best value in each column is shown in bold.}
\label{tab:dataset1_results}
\resizebox{\textwidth}{!}{
\begin{tabular}{lccccccc}
\hline
Method & Accuracy & Precision & Recall & F1-score & Specificity & ROC-AUC & PR-AUC\\
\hline
Centralized (RF)        & 0.9113 & \textbf{0.9135} & 0.8953 & 0.9043 & \textbf{0.9253} & 0.9670 & 0.9604\\
Centralized (LR)        & \textbf{0.9210} & 0.9130 & 0.9189 & \textbf{0.9159} & 0.9229 & \textbf{0.9756} & \textbf{0.9709}\\
Centralized (XGB)       & 0.9152 & 0.9074 & 0.9121 & 0.9097 & 0.9179 & 0.9725 & 0.9696\\
Centralized (GB)        & 0.9092 & 0.9098 & 0.8949 & 0.9023 & 0.9218 & 0.9701 & 0.9663\\
Centralized (AdaBoost)  & 0.9109 & 0.9068 & 0.9025 & 0.9046 & 0.9183 & 0.9711 & 0.9683\\
Centralized (ChurnNet)  & 0.9157 & 0.8968 & \textbf{0.9269} & 0.9116 & 0.9060 & 0.9735 & 0.9706\\
\hline
Local (LR)              & \textbf{0.9309} & \textbf{0.9272} & 0.9253 & \textbf{0.9262} & \textbf{0.9359} & \textbf{0.9801} & \textbf{0.9769}\\
\hline
FedAvg                  & 0.9169 & 0.8977 & \textbf{0.9285} & 0.9128 & 0.9066 & 0.9740 & 0.9696\\
FedProx                 & 0.9195 & 0.9063 & 0.9237 & \textbf{0.9150} & 0.9158 & \textbf{0.9750} & \textbf{0.9714}\\
DP-FedAvg               & \textbf{0.9193} & \textbf{0.9125} & 0.9157 & 0.9141 & \textbf{0.9225} & 0.9727 & 0.9676\\
Proposed DP-FedProx     & 0.9161 & 0.8981 & 0.9261 & 0.9119 & 0.9073 & 0.9735 & 0.9692\\
\hline
\end{tabular}}
\end{table*}

Table~\ref{tab:dataset1_results} presents the experimental results of the centralized, local, federated, and differentially private federated learning models on Dataset-1. Among the centralized models, Logistic Regression achieved the best overall performance with an accuracy of 92.10\%, an F1-score of 91.59\%, a ROC-AUC of 97.56\%, and a PR-AUC of 97.09\%. The ChurnNet obtained a comparable performance while achieving the highest recall (92.69\%) among the centralized approaches. This result shows the potential of ChurnNet model to correctly identify the churn customers. Most of the cases, the local (siloed) Logistic Regression model achieved the highest performance. As among the centralize model LR model shows better overall performance, we trained only the LR classifier as a local. This local model shows best overall performance because all four local model trained and tested on its own data where the individual records maintain a similar pattern.  This models trained on client-specific data and the local approach cannot exploit knowledge from other clients and therefore cannot be used for collaborative learning across multiple telecom operators. Among the federated learning methods, FedProx consistently outperformed FedAvg on most evaluation metrics. Specifically, FedProx improved the accuracy from 91.69\% to 91.95\%, the F1-score from 91.28\% to 91.50\%. These shows the effectiveness of FedProx approach on non-IID data distribution. The prediction results of he differential privacy based FedAvg and FedProx models show that introducing differential privacy does not reduce models performance much.  The proposed DP-FedProx model achieved an accuracy of 91.61\%, an F1-score of 91.19\%, a ROC-AUC of 97.35\%, and a PR-AUC of 96.92\%, which are comparable with non-DP federated models.

%Overall, the results indicate that FedProx provides better performance than the standard FedAvg algorithm under heterogeneous client data distributions, while the proposed DP-FedProx framework preserves most of the predictive performance despite the addition of differential privacy. These findings demonstrate that privacy-preserving federated learning is a practical solution for collaborative customer churn prediction without requiring telecom operators to share their sensitive customer data.

\begin{figure}[!htb]
\begin{center}
\includegraphics[height=330px,width=400px]{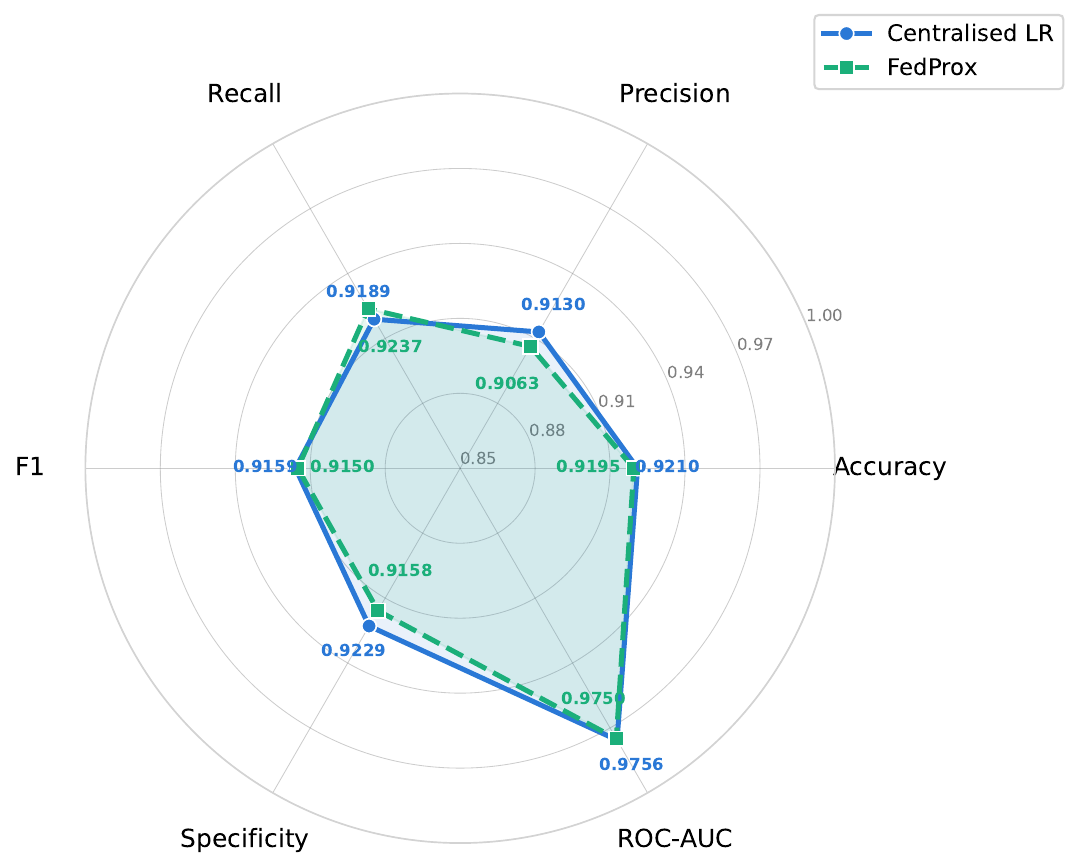}
\caption{Multi-metric radar: Centralised LR vs FedProx (Dataset-1)}
\label{fig:radar_dataset_1}
\end{center}
\end{figure}

The radar chart in Figure~\ref{fig:radar_dataset_1} compares the best centralized model (Centralized LR) with the best federated model (FedProx) across six evaluation metrics. PR-AUC is excluded for better visualization. Both models show very similar performance, indicating that FedProx achieves comparable results to centralized learning. While Centralized LR provides slightly higher Precision and Specificity, FedProx achieves higher Recall, making it more effective at identifying customers who are likely to churn. Those results demonstrate that FedProx maintains strong predictive performance while enabling privacy-preserving collaborative learning without sharing raw customer data.

%The multi-metric radar chart in Figure~\ref{fig:radar} offers a holistic comparison of Centralised LR, FedProx, and the Local LR upper bound across six performance dimensions. Local LR occupies the outermost boundary on every axis, confirming its position as the empirical ceiling on this dataset; however, as noted earlier, this configuration does not constitute a deployable privacy-preserving solution. Centralised LR and FedProx trace broadly similar polygons, with Centralised LR holding a modest edge in Precision (0.9130 vs.\ 0.9063) and Specificity (0.9229 vs.\ 0.9158), while FedProx surpasses it on Recall (0.9237 vs.\ 0.9189), indicating that federated training with proximal regularisation captures a slightly broader positive-class signal at the cost of a marginal increase in false positives. The near-congruence of the two polygons on the Accuracy, F1, and ROC-AUC axes---where the gap between Centralised LR and FedProx does not exceed 0.0015 on any of these metrics---demonstrates that the federated approach incurs negligible performance degradation relative to its centralised counterpart. Taken together, these observations suggest that FedProx achieves a compelling balance across all six metrics, making it a strong candidate for privacy-sensitive deployments where centralised data aggregation is not permissible.

\begin{table}[!ht]
\centering
\caption{Summary of performance improvements among the federated learning methods on Dataset-1.}
\label{tab:improvement_summary}
\begin{tabular}{llc}
\hline
Comparison & Better Method & Improvement \\
\hline
FedProx vs. FedAvg      & FedProx      & +0.26 pp Accuracy \\
FedProx vs. FedAvg      & FedProx      & +0.22 pp F1-score \\
\hline
DP-FedAvg vs. FedAvg    & DP-FedAvg    & +0.24 pp Accuracy \\
DP-FedAvg vs. FedAvg    & DP-FedAvg    & +0.13 pp F1-score \\
\hline
FedProx vs. DP-FedProx  & FedProx      & +0.34 pp Accuracy \\
FedProx vs. DP-FedProx  & FedProx      & +0.31 pp F1-score \\
\hline
\end{tabular}
\end{table}

Table~\ref{tab:improvement_summary} presents the performance differences among the federated learning models in term of accuracy and F1-score. Compared with the standard FedAvg algorithm, FedProx consistently achieved better predictive performance. The table shows that the FedProx improving  accuracy by 0.26\% and  F1-score by 0.22\%. When DP applied, FedAvg did not degrade performance. Instead, FedAvg+DP slightly improved. In contrast, the proposed DP-FedProx framework showed only a marginal reduction in performance compared with the non-DP FedProx model with decreases of 0.34\% in accuracy and 0.31\% in F1-score. These experimental results indicate that the proposed framework preserves most of the predictive capability while providing formal differential privacy guarantees.

%These improvements demonstrate the effectiveness of the proximal regularization term in addressing the challenges posed by heterogeneous client data.

\subsection{Results on Dataset-2} \label{sec:exp_dataset2}

\begin{table*}[!ht]
\centering
\caption{Performance comparison of centralized, local, federated, and differentially private federated learning models on Dataset-2. The best value within the group in each column is shown in bold.} 
\label{tab:dataset2_results}
\resizebox{\textwidth}{!}{
\begin{tabular}{lccccccc}
\hline
Method & Accuracy & Precision & Recall & F1-score & Specificity & ROC-AUC & PR-AUC\\
\hline
Centralized (RF)        & 0.8266 & 0.7110 & 0.5856 & 0.6422 & 0.9138 & 0.8657 & 0.6969\\
Centralized (LR)        & \textbf{0.8351} & \textbf{0.7247} & 0.6123 & \textbf{0.6638} & \textbf{0.9158} & 0.8737 & \textbf{0.7337}\\
Centralized (XGB)       & 0.8223 & 0.6742 & 0.6417 & 0.6575 & 0.8877 & 0.8581 & 0.6801\\
Centralized (GB)        & 0.8244 & 0.6851 & 0.6283 & 0.6555 & 0.8955 & 0.8817 & 0.7304\\
Centralized (AdaBoost)  & 0.8259 & 0.6903 & 0.6257 & 0.6564 & 0.8984 & \textbf{0.8836} & 0.7336\\
Centralized (ChurnNet)  & 0.8124 & 0.6478 & \textbf{0.6444} & 0.6461 & 0.8732 & 0.8446 & 0.6537\\
\hline
Local (LR)              & \textbf{0.8429} & \textbf{0.7460} & 0.6203 & \textbf{0.6774} & \textbf{0.9235} & \textbf{0.8958} & \textbf{0.7578}\\
\hline
FedAvg                  & 0.7825 & 0.6133 & 0.4920 & 0.5460 & 0.8877 & 0.8176 & 0.6125\\
FedProx                 & 0.7903 & 0.6179 & 0.5535 & 0.5839 & 0.8761 & 0.8279 & 0.6212\\
DP-FedAvg               & \textbf{0.8031} & \textbf{0.6559} & 0.5455 & \textbf{0.5956} & \textbf{0.8964} & \textbf{0.8350} & \textbf{0.6591}\\
Proposed DP-FedProx     & 0.8003 & 0.6458 & 0.5508 & 0.5945 & 0.8906 & 0.8272 & 0.6322\\
\hline
\end{tabular}}
\end{table*}

Table~\ref{tab:dataset2_results} presents the experimental results on dataset-2 across the all evaluation metrics for the centralized, local, federated, and differentially private federated learning models. Among the centralized models, Logistic Regression achieved the best overall performance, with an accuracy of 83.51\%, an F1-score of 66.38. ChurnNet obtained the highest recall (64.44\%) among the centralized approaches, indicating its better ability to identify churn customers. The Local (LR) model achieved the highest overall performance on Dataset-2. However, as we discussed earlier, each local model is trained only on its own data, which have more similar patterns and this approach cannot support collaborative learning across multiple telecom operators. 

Among the federated learning methods without differential privacy, FedProx shows better perdiction results than FedAvg. When differential privacy applied with federated learning DP-FedAvg achieved better perfromance than DP-FedProx model. 

%The introduction of differential privacy further improved the performance of FedAvg. FedAvg+DP achieved the best performance among all federated models, with an accuracy of 80.31\%, an F1-score of 59.56\%, a ROC-AUC of 83.50\%, and a PR-AUC of 65.91\%. Compared with FedProx, the proposed DP-FedProx framework showed only a slight reduction in Accuracy (0.36 percentage points) and PR-AUC (1.10 percentage points), while improving Recall by 1.94 percentage points. This indicates that the proposed framework maintains competitive predictive performance while providing formal differential privacy guarantees.

\begin{figure}[!htb]
\begin{center}
\includegraphics[height=330px,width=400px]{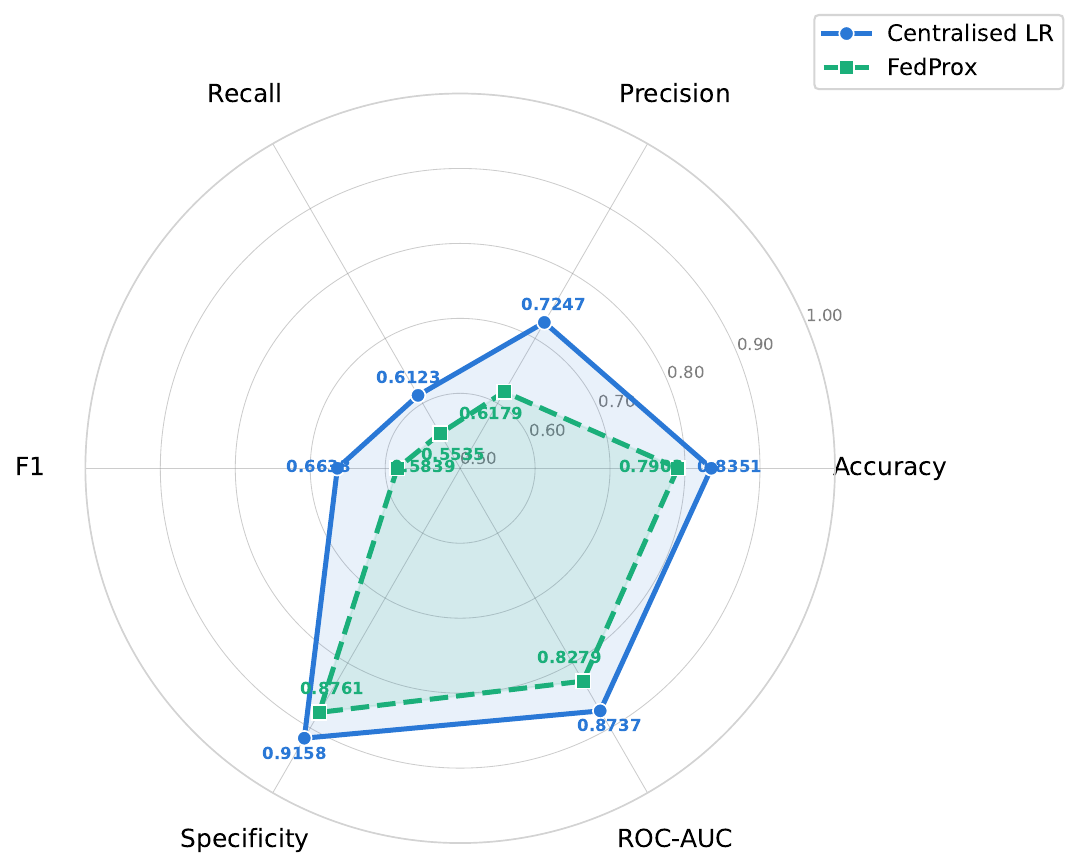}
\caption{Multi-metric radar: Centralised LR vs FedProx (Dataset-2)}
\label{fig:radar_dataset_2}
\end{center}
\end{figure}

To highlight the performance difference between the best centralized LR and the proposed FedProx federated model, we draw a radar chart presented in Figure~\ref{fig:radar_dataset_2}. It shows more wider performance gap between the two models than that observed on Dataset-1. Centralized LR outperforms the FedProx on every metric. Unlike Dataset-1, where the two radar shapes were almost the same, the FedProx shape in Dataset-2 is much smaller, showing a larger drop in performance. This is likely because Dataset-2 is more difficult and has a higher level of non-IID data across clients. These results show that the performance loss in federated learning depends on the characteristics of the dataset.

\begin{table}[!ht]
\centering
\caption{Summary of performance improvements among the federated learning methods on Dataset-2.}
\label{tab:improvement_summary_dataset2}
\begin{tabular}{llc}
\hline
Comparison & Better Method & Improvement \\
\hline
FedProx vs. FedAvg      & FedProx      & +0.78 pp Accuracy \\
FedProx vs. FedAvg      & FedProx      & +3.79 pp F1-score \\
\hline
DP-FedAvg vs. FedAvg    & DP-FedAvg    & +2.06 pp Accuracy \\
DP-FedAvg vs. FedAvg    & DP-FedAvg    & +4.96 pp F1-score \\
\hline
FedProx vs. DP-FedProx  & DP-FedProx      & +1 pp Accuracy \\
FedProx- vs. DP-FedProx  & DP-FedProx & +1.06 pp F1-score \\
\hline
\end{tabular}
\end{table}

Table~\ref{tab:improvement_summary_dataset2} summarizes the performance improvements of the federated learning methods on Dataset-2. Compared with FedAvg, FedProx improves the Accuracy by 0.78\% and the F1-score by 3.79\%. Adding differential privacy to FedAvg also improves its performance, increasing Accuracy by 2.06\% and F1-score by 4.96\% over the standard FedAvg model. When comparing FedProx with DP-FedProx, the DP-FedProx model achieves slightly better results, with improvements of 1\% in Accuracy and 1.06\% in F1-score.These results show that both the FedProx algorithm and differential privacy can improve the performance of federated learning on Dataset-2.

\section{Comparison with Other Studies} \label{sec:comp_other_study}
To the best of our knowledge, we found only two previous research articles~\cite{Huh2024,Krishnan2024} that focus on federated learning for customer churn prediction in the telecommunications domain. However, in~\cite{Huh2024}, the proposed framework does not fully preserve the core privacy principle of federated learning, as it stores raw customer data in a centralized global repository before model training. On the other hand, the study by~\cite{Krishnan2024} did not incorporate differential privacy and was evaluated using a private dataset. Since the experimental setups of these studies differ significantly from ours, a direct comparison of prediction performance is not appropriate. Instead, Table~\ref{tab:comparison_related_work} compares the key characteristics of our proposed framework with those of the two existing studies~\cite{Huh2024,Krishnan2024}.

\begin{table}[!ht]
\centering
\caption{Comparison between the proposed framework and the related work.}
\label{tab:comparison_related_work}
\begin{tabular}{p{4cm}p{3cm}p{3cm}p{4cm}}
\hline
\textbf{Feature} & \textbf{Huh and Lee~\cite{Huh2024}} &\textbf{Krishnan~\cite{Krishnan2024}} & \textbf{Proposed Framework} \\
\hline
Application & Telecom churn prediction &  Telecom churn prediction & Telecom churn prediction \\
Federated algorithm & FedAvg & FedAvg & FedAvg, DP-FedAvg, FedProx, DP-FedProx \\
Statistical heterogeneity (non-IID data) & Limited & Limited & Explicitly addressed using FedProx\\
Differential privacy & Not applied & Not applied & Integrated into FedProx \\
Raw data handling & Centralized before client distribution & Data remain at local clients & Data remain at local clients \\
Datasets & One telecom dataset (marges the two datasets) & One dataset & Two public telecom datasets \\
Performance evaluation & Limited comparison & Limited comparison  & Comprehensive comparison with centralized and federated models \\
Best Accuracy & 86.50\%  & 84.40\%  &  91.93\% \\
\hline
\end{tabular}
\end{table}

\section{Data Privacy Guarantee}
\label{sec:privacy_guarantee}
The proposed framework protects data privacy in two ways. First, we train a model using federated learning which allows multiple telecom operators or  branches to collaboratively train a global model without exchanging their raw information. This approach reduces the risk of data leakage during the training process and helps telecom operators comply with data privacy regulations. 
Second, the proposed framework applies Differentially Private Stochastic Gradient Descent (DP-SGD) during local model training. In DP-SGD, the Gaussian mechanism adds noise to the clipped gradients before the local model parameters are updated. After local training, the model updates are sent to the central server. This makes it difficult for an adversary to recover sensitive customer information from the shared model updates or the final trained model. In this study, we set the privacy budget to $\varepsilon=1$ and $\delta=10^{-5}$ for both DP-FedAvg and DP-FedProx. Several previous studies recommend that a privacy budget of $\varepsilon \leq 10$  can provide meaningful differential privacy guarantees \cite{JDK_2025_PPCCP, Ponomareva2023,  xie2018differentially, Huang2023}. Therefore, our choice of $\varepsilon=1$ provides a strong level of privacy protection. The experimental results show that the proposed DP-FedProx framework achieves prediction performance close to the without DP FedProx model.

%Bridging the Gap: DP-FedProx, DP-FedAvg, and the Central (LR) Baseline
\section{Bridging the Gap: Why DP based FedProx and FedAvg Achieve Comparable Performance} 
\label{sec:DP-effect}
This research shows that DP-FedAvg and DP-FedProx achieve comparable performance, with only a small gap between them. This indicates that both the privacy noise introduced by DP and the client heterogeneity have a limited impact in our setting. We identify two contributing reasons: mild client heterogeneity and small proximal correction compared with DP noise.

FedProx is most useful when the data distributions of different clients are very different. In our experiments, clients are divided according to customer tenure. The four groups are $\leq 12$, $13$--$24$, $25$--$48$, and $>48$ months. Although this creates a non-IID setting, customers within each group have similar churn related characteristics. As a result, the difference between local and global objectives is relatively small. Since client drift is not severe, there is limited room for the proximal term in FedProx to provide additional improvement, either with or without DP.

FedProx uses the proximal term $\frac{\mu}{2}\lVert w_k-w_{\text{global}}\rVert^2$ to reduce the difference between the local and global models. In this study, we set $\mu=0.01$. With this value of $\mu$, the effect of this term on the gradient is relatively small. In DP-SGD, gradients are clipped to a maximum norm of $C=1.0$, and random noise is added to protect privacy. The standard deviation of this noise is proportional to $\sigma C/B$. When the DP noise is similar to or larger than the correction from the proximal term, the effect of the proximal correction can be hidden by the noise. We use a small value of $\mu$ because our tenure based client partition already produces mild client heterogeneity. A larger value of $\mu$ could over regularize the local model updates, especially when the updates already contain DP noise. In this case, the proximal term could introduce additional distortion instead of providing more useful correction. Therefore, we do not consider the small value of $\mu$ to be a limitation. Instead, it shows that DP-FedProx can perform similarly to DP-FedAvg when client heterogeneity is mild, while still providing stronger correction.

\subsection{Variable Importance via SHAP Analysis}
\label{subsec:shap-analysis}

To quantify which individual variables influence churn prediction, we computed SHAP (SHapley Additive exPlanations) values on two versions of the trained ChurnNet model for each dataset: (i) the \emph{centralized} model, trained on the full, unpartitioned dataset without federated aggregation or differential privacy, which characterizes the underlying churn-generating process independent of the training mechanism; and (ii) the proposed FedProx+DP global model ($\epsilon=1.0$), which characterizes what the actual deployed, privacy-preserving model relies on. SHAP values were estimated using \texttt{shap.GradientExplainer} on the preprocessed feature space (the same representation used for model training), with a background reference set of 200 samples and a held-out explanation set.

\subsubsection{SHAP Analysis for Centralized Model}
\label{subsubsec:shap-centralized}

\begin{figure}[h!]
    \centering
    \begin{subfigure}[b]{0.48\linewidth}
        \centering
        \includegraphics[width=\linewidth]{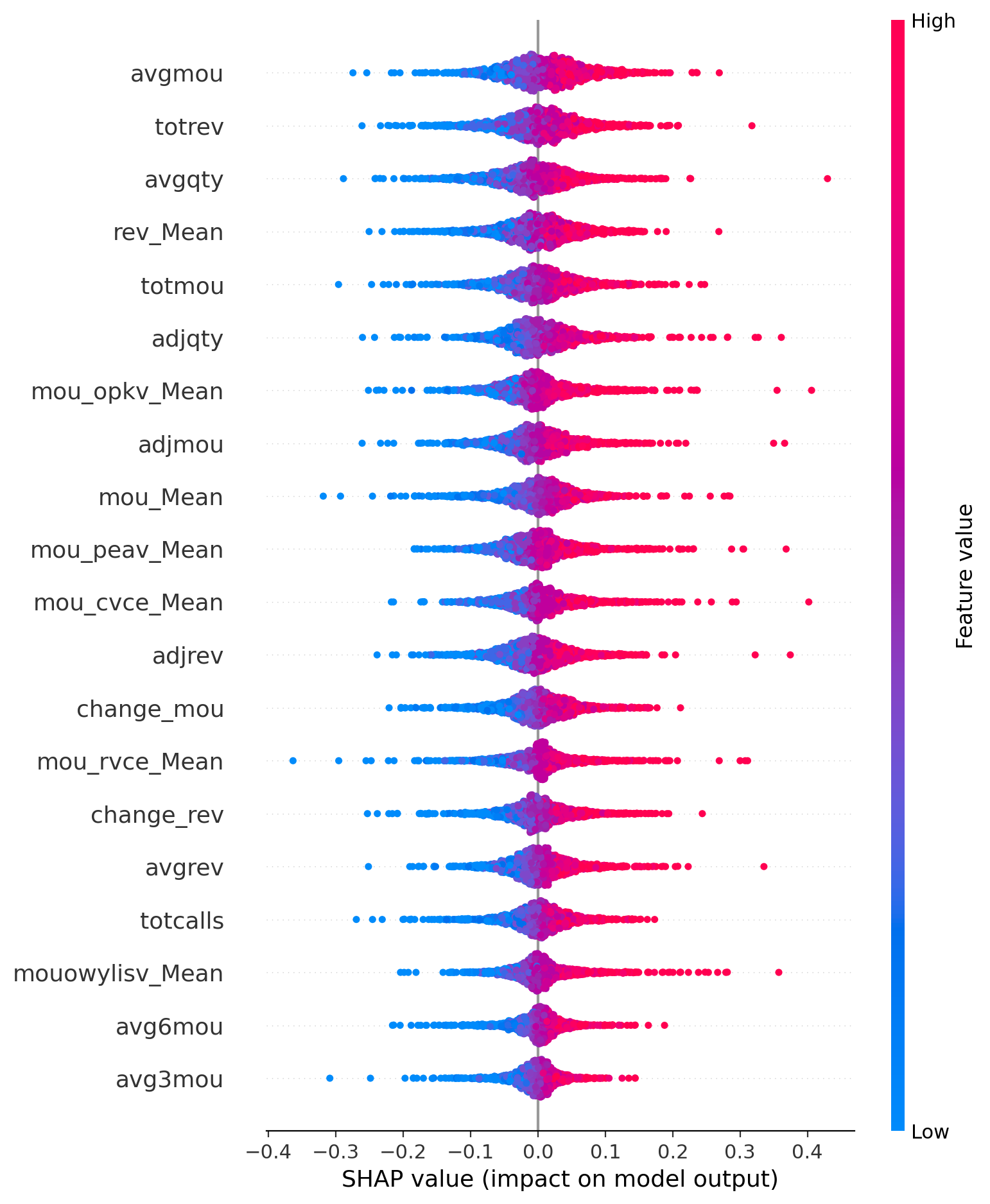}
        \caption{Dataset-1}
        \label{fig:shap-beeswarm-d1}
    \end{subfigure}
    \hfill
    \begin{subfigure}[b]{0.48\linewidth}
        \centering
        \includegraphics[width=\linewidth]{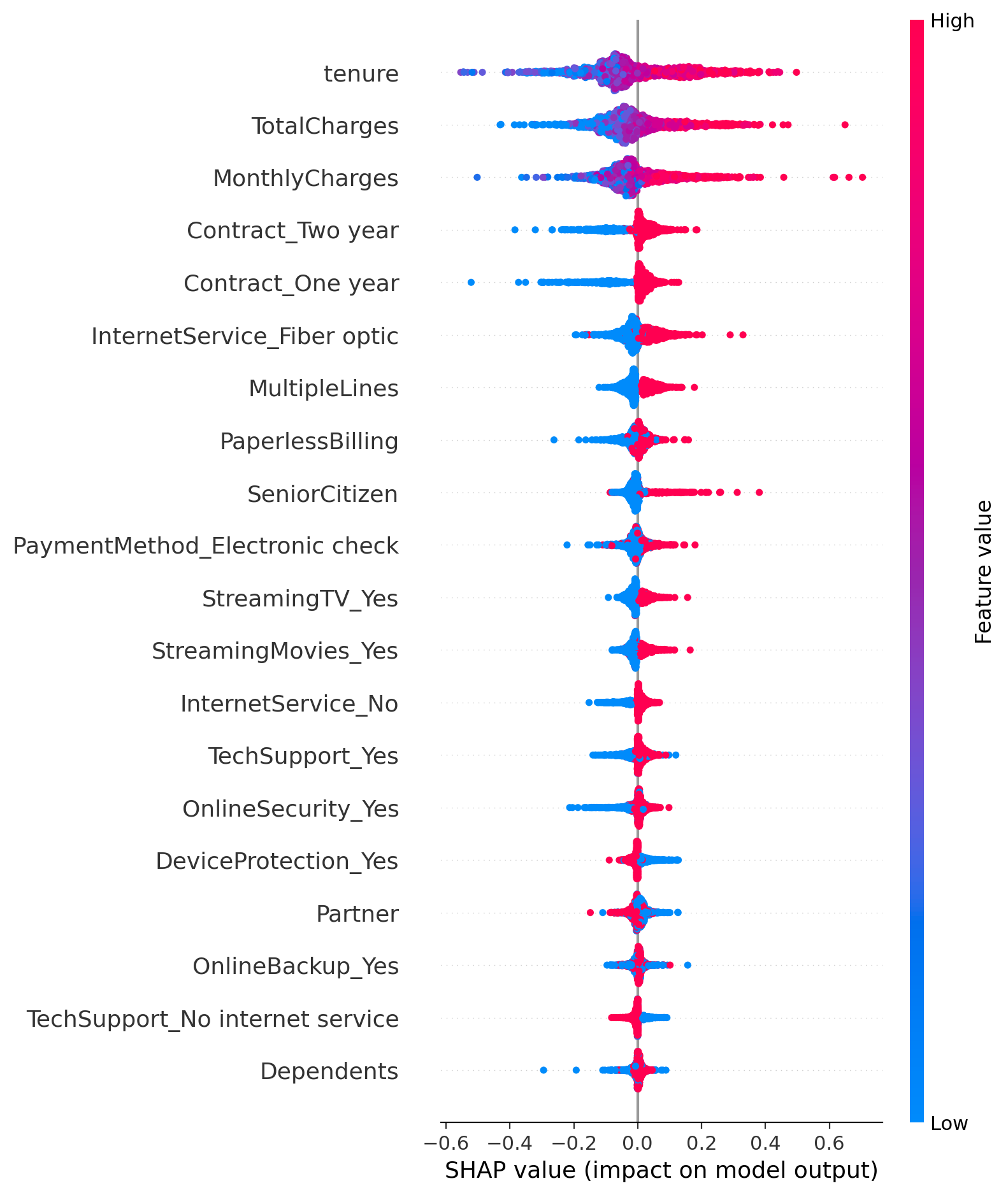}
        \caption{Dataset-2}
        \label{fig:shap-beeswarm-d2}
    \end{subfigure}
    \caption{SHAP beeswarm plots for the centralized ChurnNet model on (a) Dataset-1 and (b) Dataset-2. Each point represents one customer; horizontal position indicates the SHAP value (impact on predicted churn probability) and color indicates the feature's value for that customer.}
    \label{fig:shap-beeswarm-central}
\end{figure}

Table~\ref{table:shap-dataset1} and Table~\ref{table:shap-dataset2} report the top-10 features ranked by mean absolute SHAP value for the centralized ChurnNet model on Dataset-1 and Dataset-2, respectively. Figure~\ref{fig:shap-beeswarm-central} shows the corresponding beeswarm plots, which additionally indicate the \emph{direction} of each feature's effect on predicted churn probability.

\begin{table}[h!]
\caption{Top-10 features by mean $|\text{SHAP}|$ value - Dataset-1, centralized ChurnNet}
\label{table:shap-dataset1}
\begin{center}
\begin{tabular}{ c l c }
 \hline
 Rank & Feature & Mean $|\text{SHAP}|$ \\
 \hline
 1  & avgmou        & 0.0439 \\
 2  & totrev        & 0.0428 \\
 3  & avgqty        & 0.0426 \\
 4  & rev\_Mean     & 0.0398 \\
 5  & totmou        & 0.0397 \\
 6  & adjqty        & 0.0383 \\
 7  & mou\_opkv\_Mean & 0.0379 \\
 8  & adjmou        & 0.0374 \\
 9  & mou\_Mean     & 0.0371 \\
 10 & mou\_peav\_Mean & 0.0371 \\
 \hline
\end{tabular}
\end{center}
\end{table}

\begin{table}[h!]
\caption{Top-10 features by mean $|\text{SHAP}|$ value - Dataset-2, centralized ChurnNet}
\label{table:shap-dataset2}
\begin{center}
\begin{tabular}{ c l c }
 \hline
 Rank & Feature & Mean $|\text{SHAP}|$ \\
 \hline
 1  & tenure                          & 0.1175 \\
 2  & TotalCharges                    & 0.0876 \\
 3  & MonthlyCharges                  & 0.0751 \\
 4  & Contract\_Two year              & 0.0422 \\
 5  & Contract\_One year              & 0.0400 \\
 6  & InternetService\_Fiber optic    & 0.0382 \\
 7  & MultipleLines                   & 0.0323 \\
 8  & PaperlessBilling                & 0.0231 \\
 9  & SeniorCitizen                   & 0.0217 \\
 10 & PaymentMethod\_Electronic check & 0.0213 \\
 \hline
\end{tabular}
\end{center}
\end{table}

For Dataset-1, the top-ranked features are dominated almost entirely by usage-volume variables: average and total minutes of use (\texttt{avgmou}, \texttt{totmou}, \texttt{mou\_Mean}), call/usage quantity (\texttt{avgqty}, \texttt{adjqty}), and revenue (\texttt{totrev}, \texttt{rev\_Mean}). Notably, the top-10 mean SHAP values are very close to each other (0.0371--0.0439), indicating that no single feature has a much stronger effect on the model's decisions -- churn risk is instead explained by a combination of usage-related features, likely because the large feature space (101 attributes) spreads predictive information across many related usage measures.

In contrast, for Dataset-2, \texttt{tenure} is the clear dominant driver (mean SHAP $= 0.1175$), roughly 34\% higher than the second-ranked feature and more than 5 times the importance of the tenth-ranked feature. Billing-related features, \texttt{TotalCharges} and \texttt{MonthlyCharges}, rank second and third, followed by contract-related features, \texttt{Contract\_Two year} and \texttt{Contract\_One year}.
This supports the common finding that long-term contracts reduce the risk of churn, while their absence is a risk signal. Service related and account related features also affect churn, but their effects are smaller. Compared with the flat importance pattern in Dataset-1, Dataset-2 shows a much steeper pattern. This is likely because Dataset-2 has a smaller and more focused feature set with only 21 attributes.

\subsubsection{SHAP Analysis for DP-FedProx Global Model}
\label{subsubsec:shap-fedproxdp}

\begin{figure}[h!]
    \centering
    \begin{subfigure}[b]{0.48\linewidth}
        \centering
        \includegraphics[width=\linewidth]{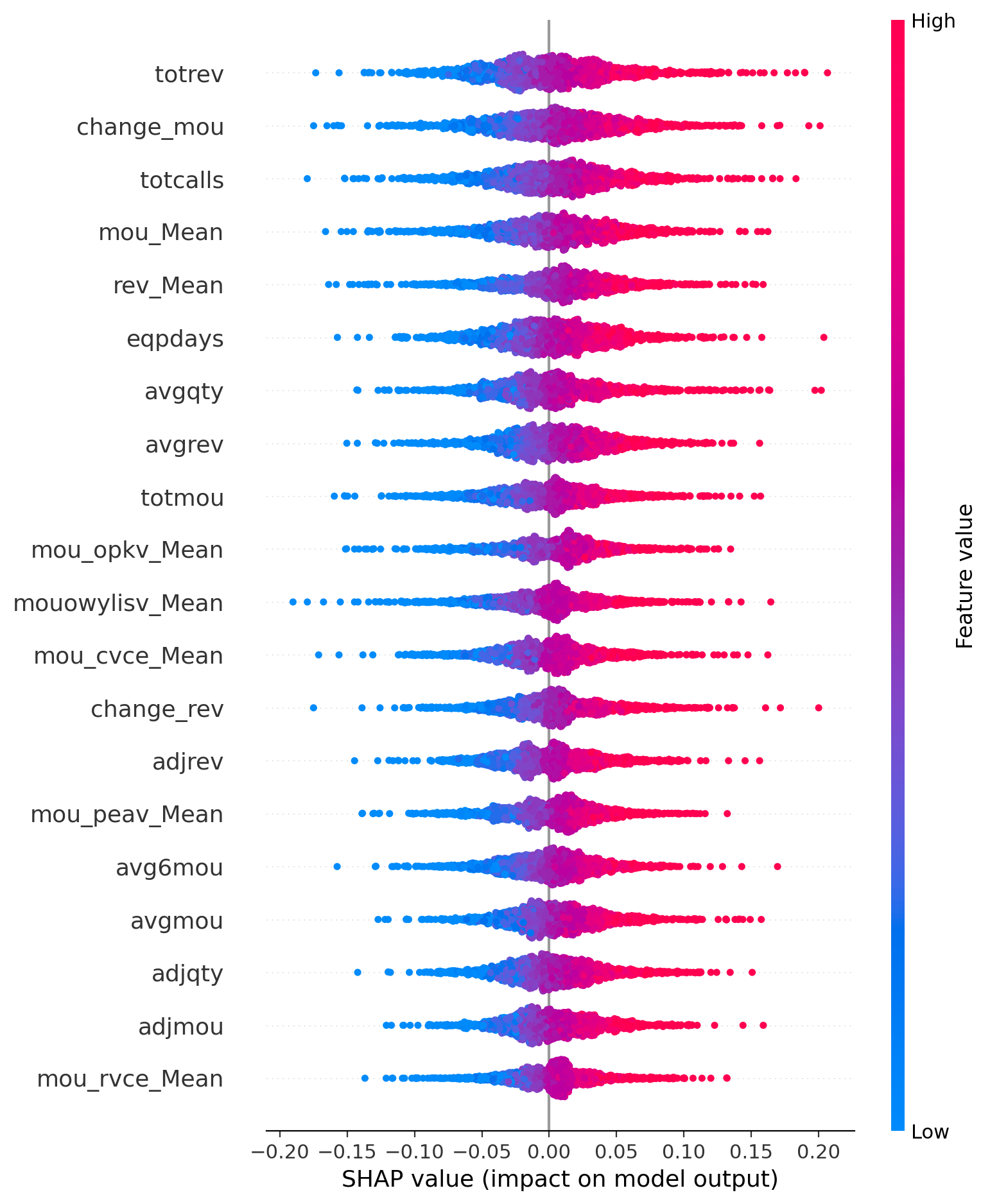}
        \caption{Dataset-1}
        \label{fig:shap-beeswarm-fedproxdp-d1}
    \end{subfigure}
    \hfill
    \begin{subfigure}[b]{0.48\linewidth}
        \centering
        \includegraphics[width=\linewidth]{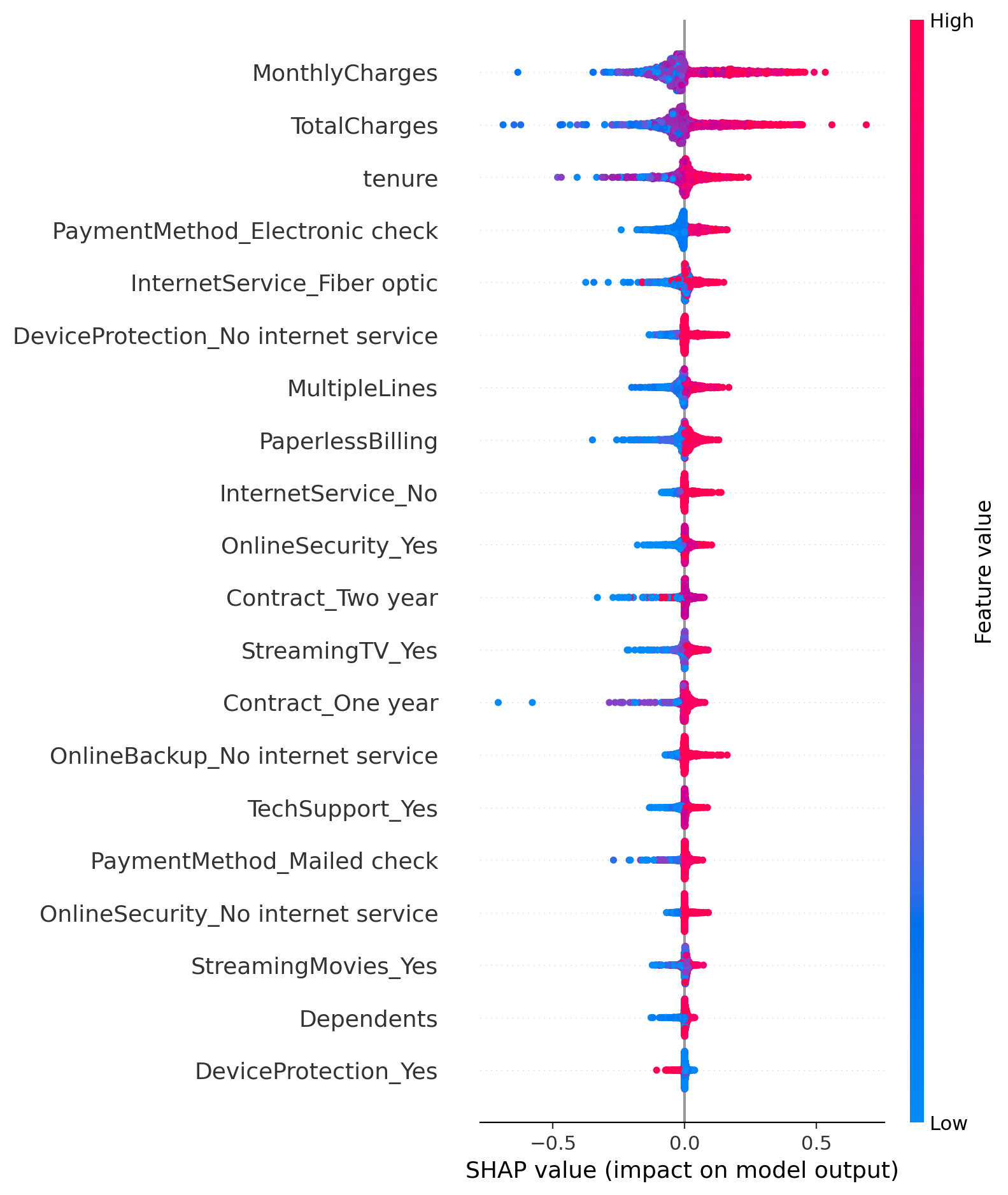}
        \caption{Dataset-2}
        \label{fig:shap-beeswarm-fedproxdp-d2}
    \end{subfigure}
    \caption{SHAP beeswarm plots for the proposed FedProx+DP global model on (a) Dataset-1 and (b) Dataset-2. Each point represents one customer; horizontal position indicates the SHAP value (impact on predicted churn probability) and color indicates the feature's value for that customer.}
    \label{fig:shap-beeswarm-fedproxdp}
\end{figure}

To assess whether the centralized-model drivers also hold their important for the model actually proposed in this work, we repeated the SHAP analysis on the trained FedProx+DP global model ($\epsilon = 1.0$). This model was created by combining the models from four tenure-based clients using federated aggregation with differentially private local training. Table~\ref{table:shap-dataset1-fedproxdp} and Table~\ref{table:shap-dataset2-fedproxdp} report the resulting top-10 features, and Figure~\ref{fig:shap-beeswarm-fedproxdp} shows the corresponding beeswarm plots.

\begin{table}[h!]
\caption{Top-10 features by mean $|\text{SHAP}|$ value - Dataset-1, proposed FedProx+DP global model ($\epsilon=1.0$)}
\label{table:shap-dataset1-fedproxdp}
\begin{center}
\begin{tabular}{ c l c }
 \hline
 Rank & Feature & Mean $|\text{SHAP}|$ \\
 \hline
 1  & totrev        & 0.0356 \\
 2  & change\_mou   & 0.0355 \\
 3  & totcalls      & 0.0343 \\
 4  & mou\_Mean     & 0.0330 \\
 5  & rev\_Mean     & 0.0316 \\
 6  & eqpdays       & 0.0315 \\
 7  & avgqty        & 0.0299 \\
 8  & avgrev        & 0.0299 \\
 9  & totmou        & 0.0295 \\
 10 & mou\_opkv\_Mean & 0.0286 \\
 \hline
\end{tabular}
\end{center}
\end{table}

\begin{table}[h!]
\caption{Top-10 features by mean $|\text{SHAP}|$ value - Dataset-2, proposed FedProx+DP global model ($\epsilon=1.0$)}
\label{table:shap-dataset2-fedproxdp}
\begin{center}
\begin{tabular}{ c l c }
 \hline
 Rank & Feature & Mean $|\text{SHAP}|$ \\
 \hline
 1  & MonthlyCharges                       & 0.0911 \\
 2  & TotalCharges                         & 0.0837 \\
 3  & tenure                               & 0.0439 \\
 4  & PaymentMethod\_Electronic check      & 0.0331 \\
 5  & InternetService\_Fiber optic         & 0.0276 \\
 6  & DeviceProtection\_No                 & 0.0251 \\
 7  & MultipleLines                        & 0.0244 \\
 8  & PaperlessBilling                     & 0.0244 \\
 9  & InternetService\_No                  & 0.0173 \\
 10 & OnlineSecurity\_Yes                  & 0.0162 \\
 \hline
\end{tabular}
\end{center}
\end{table}

For Dataset-1, the FedProx+DP model's top features remain dominated by usage and revenue variables, similar to the centralized model. The feature importance is also relatively flat, with mean SHAP values ranging from 0.0286 to 0.0356, consistent with the centralized model. However, the ranking of individual features changes. Features such as \texttt{eqpdays} (days since the last equipment upgrade), \texttt{change\_mou} (change in minutes of use), \texttt{totcalls}, and \texttt{avgrev} appear in the top 10. They replace \texttt{adjqty}, \texttt{adjmou}, and \texttt{mou\_peav\_Mean}, which were in the top 10 of the centralized model. This suggests that federated aggregation and DP noise cause the global model to rely more on changes in usage and equipment-related information, rather than only on total usage levels. However, the overall importance of usage and revenue features remains similar.

For Dataset-2, the change of the importance features is more noticeable: billing variables (\texttt{MonthlyCharges}, \texttt{TotalCharges}) become the two strongest predictors. In the centralized model, \texttt{tenure} was the most important feature, but in the FedProx+DP model, it falls to third place. The contract-related features, \texttt{Contract\_Two year} and \texttt{Contract\_One year}, which were important in the centralized model, are no longer in the top 10. Instead, service related features such as \texttt{DeviceProtection\_No}, \texttt{InternetService\_No}, and \texttt{OnlineSecurity\_Yes} enter the top 10. These results indicate that federated learning across clients with different tenure levels, together with DP-SGD noise, changes the global model's feature importance. The model relies less on the single dominant feature, \texttt{tenure}, and more balanced set of billing and service-related features.

The SHAP analysis for the centralized and FedProx+DP models shows that the strongest predictors of churn can shift when the training methodology changes, and we observe this pattern in both datasets. It is an important finding that shows how federated aggregation and differential privacy can change the features on which the global model relies. In other words, the features that strongly explain churn in the original data are not always the same features that the privacy-protected federated model relies on most. This difference provides useful information about how DP-SGD affects the balance between model performance, privacy, and interpretability in a federated learning setting.

%\textbf{Implication for the non-IID partitioning:} The centralized-model analysis in Section~\ref{subsubsec:shap-centralized} shows that \texttt{tenure} is a strong predictor of churn in Dataset-2. In Dataset-1, features related to accumulated usage, which are also linked to tenure, are among the most important features. This supports the use of tenure as a meaningful and non-arbitrary basis for creating non-IID client groups, as discussed in Section~\ref{subsec:data-partitioning}. However, the FedProx+DP results show that tenure becomes less important in the final global model compared with the centralized model. This change is especially clear in Dataset-2, where \texttt{tenure} moves from first place to third place. We do not consider this a reason to reject tenure-based client partitioning. Instead, it is an important finding that shows how federated aggregation and differential privacy can change the features on which the global model relies. In other words, the features that strongly explain churn in the original data are not always the same features that the privacy-protected federated model relies on most. This difference provides useful information about how DP-SGD affects the balance between model performance, privacy, and interpretability in a federated learning setting.

\section{Discussion} \label{sec:discussion}

This study evaluated several Federated Learning approaches with differential privay and without differential privacy  for customer churn prediction in the telecom industry. We also trained several centralized and local models. The results from both datasets show that federated learning can achieve good prediction performance while protecting customer privacy.

For Dataset-1, among the centralized models, Logistic Regression shows the best overall prediction performance, while ChurnNet achieved the highest Recall, meaning it can identify more churn customers.The local LR model shows overall better performance than other. This is beacause, each local model trained and tested on its own data where individual records are follow a similar pattern. Among the federated learning methods, FedProx performed better than FedAvg, showing that it can better handle non-IID client data. The proposed DP-FedProx model achieved results very close to FedProx, with only a small decrease in Accuracy and F1-score. 
For both DP-FedProx and DP-FedAvg, we set the privacy budget to $\varepsilon=1$. According to previous studies, a privacy budget of $\varepsilon \leq 10$ is generally recommended to provide meaningful differential privacy guarantees~\cite{Ponomareva2023,JDK_2025_PPCCP, xie2018differentially,Huang2023}. Therefore, our choice of $\varepsilon=1$ provides a strong level of privacy protection. Even with this strong privacy guarantee, the proposed framework achieved prediction performance very close to the non-private FedProx model.

Almost similar results were also observed for Dataset-2. The Logistic Regression was the best centralized model. FedProx consistently performed better than FedAvg, showing its ability to handle heterogeneous client data. The results also show that adding differential privacy caused only a small change in prediction performance. In fact, FedAvg+DP achieved the highest Accuracy and F1-score among the federated models on Dataset-2. The proposed DP-FedProx model also produced competitive results and achieved a higher Recall than the standard FedProx model.

From the radar charts on dataset-1, we observed that the federated models have performance patterns very close to the best centralized model Logistic Regression. But for the dataset-2, federated approach show lower performance than the best centralized model. A likely reason is that Dataset-2 presents a more challenging prediction task, as both centralized and federated models achieved lower performance. The heterogeneous (non-IID) distribution of client data may have had a stronger impact on federated training which make more difficult for the global model to converge. These findings suggest that the performance of federated learning depends on the characteristics of the dataset and the level of data heterogeneity across clients.

The bridging the gap analysis indicates that the small difference between DP-FedAvg and DP-FedProx is mainly due to two factors: mild client heterogeneity under our tenure-based partitioning, and the proximal correction is small compared with the noise added by DP-SGD. This suggests that when client groups are not very different, the proximal term may provide only limited practical benefits after adding privacy noise. However, it may be more useful when the client data are more heterogeneous.

The SHAP analysis also shows that the ranking of the most important churn features changes between the centralized and FedProx+DP models for both datasets. This means that federated aggregation and DP-SGD do more than simply add noise to the model. They can also change which features the global model relies on most. This has an important practical implication for federated churn prediction: feature-importance results from a centralized model may not be the same as those of the privacy-preserving federated model used in practice. Therefore, interpretability should ideally be evaluated on the final deployed federated model itself.

Overall, the experimental results show that the proposed DP-FedProx framework provides a good balance between prediction performance and privacy protection. It allows multiple telecom operators to train a shared customer churn prediction model without exchanging raw customer data. The results also demonstrate that strong privacy protection ($\varepsilon=1$) can be achieved with only a small reduction in prediction performance. Therefore, the proposed framework is suitable for real-world telecom networks where customer privacy and data confidentiality are important.

\section{Conclusions} \label{sec:conclusions}
Maintaining privacy of personal data is a difficult task. In this research, we used both Differential Privacy and Federated learning to mitigate privacy concern. We evaluated several approaches and finally, we proposed a Differentially Private FedProx framework for customer churn prediction in heterogeneous federated telecom networks. The framework combines FedProx with Differentially Private Stochastic Gradient Descent (DP-SGD) to protect customer privacy. This framework allows multiple telecom operators to train a shared prediction model. It also considers the non-IID data problem by partitioning clients based on customer tenure. The SHAP analysis shows that the FedProx-DP method changes the impact of features on churn prediction. 

The experimental results on two public availabe telecom churn datasets show that the proposed framework achieves competitive prediction performance and preserves data privacy. The comparative analysis indicates that FedProx consistently produced better results than the standard FedAvg algorithm under heterogeneous client data distributions. Introducing differential privacy with federated learning caused only a small decrease in performance.  These results show that strong privacy protection can be achieved without significantly reducing prediction accuracy. The proposed framework provides a practical solution for collaborative customer churn prediction in situations where telecom operators cannot share raw customer data because of privacy regulations, security requirements, or business confidentiality. Therefore, it is a promising approach for privacy-preserving machine learning in real-world federated telecom networks. In future work, we plan to investigate adaptive privacy budget allocation, personalized federated learning, and transformer-based federated models to further improve prediction performance and privacy protection.

\section*{Declaration of Generative AI Use}
During the preparation of this work the author used ChatGPT and Claude in order to assist with language editing, grammar correction, and improving the readability of the manuscript. After using these tools/services, the author reviewed and edited the content as needed and take full responsibility for the content of the published article.

\bibliographystyle{elsarticle-num}

\bibliography{bibliography}

\end{document}